\documentclass{article}

\PassOptionsToPackage{numbers, compress}{natbib}
\usepackage[preprint]{neurips_2025}

\def\eg{\emph{e.g.}} 
\def\ie{\emph{i.e.}}

\usepackage[utf8]{inputenc} 
\usepackage[T1]{fontenc}    
\usepackage{hyperref}       
\usepackage{url}            
\usepackage{booktabs}       
\usepackage{amsfonts}       
\usepackage{nicefrac}       
\usepackage{microtype}      
\usepackage{xcolor}         

\usepackage{algorithm,algorithmic}
\usepackage{amsmath}
\usepackage{bm}
\usepackage{graphicx}
\usepackage{subcaption}
\usepackage{colortbl,pifont,lipsum,graphics}

\usepackage{multirow}
\newcommand{\cmark}{\textcolor[rgb]{0.004, 0.58, 0.122}{\ding{51}}} 
\newcommand{\xmark}{\textcolor{red}{\ding{55}}} 

\usepackage{cleveref}

\title{Evaluating and Advancing Multimodal Large Language Models in Perception Ability Lens}

\author{
Feng Chen$^{1*}$ \quad Chenhui Gou$^{2*}$ \quad  Jing Liu$^2$ \quad Yang Yang$^3$ \quad Zhaoyang Li$^4$ \quad Jiyuan Zhang$^{4\dagger}$ \\ \quad \textbf{Zhenbang Sun}$^4$ \quad \textbf{Bohan Zhuang}$^{5\dagger}$ \quad \textbf{Qi Wu}$^{1\dagger}$ \\[2mm]
 \small $^*$equal technical contribution,  $^{\dagger}$equal advising \\[2mm] 
 $^1$AIML, University of Adelaide  $^2$Monash University  \\  $^3$Australian National University $^4$Tiktok, Australia  $^5$Zhejiang University 
\vspace{-1em}
}

\begin{document}

\maketitle

\begin{figure*}[!h]
    \centering
    \begin{subfigure}[b]{0.49\textwidth}
        \centering
        \includegraphics[width=1.15\textwidth]{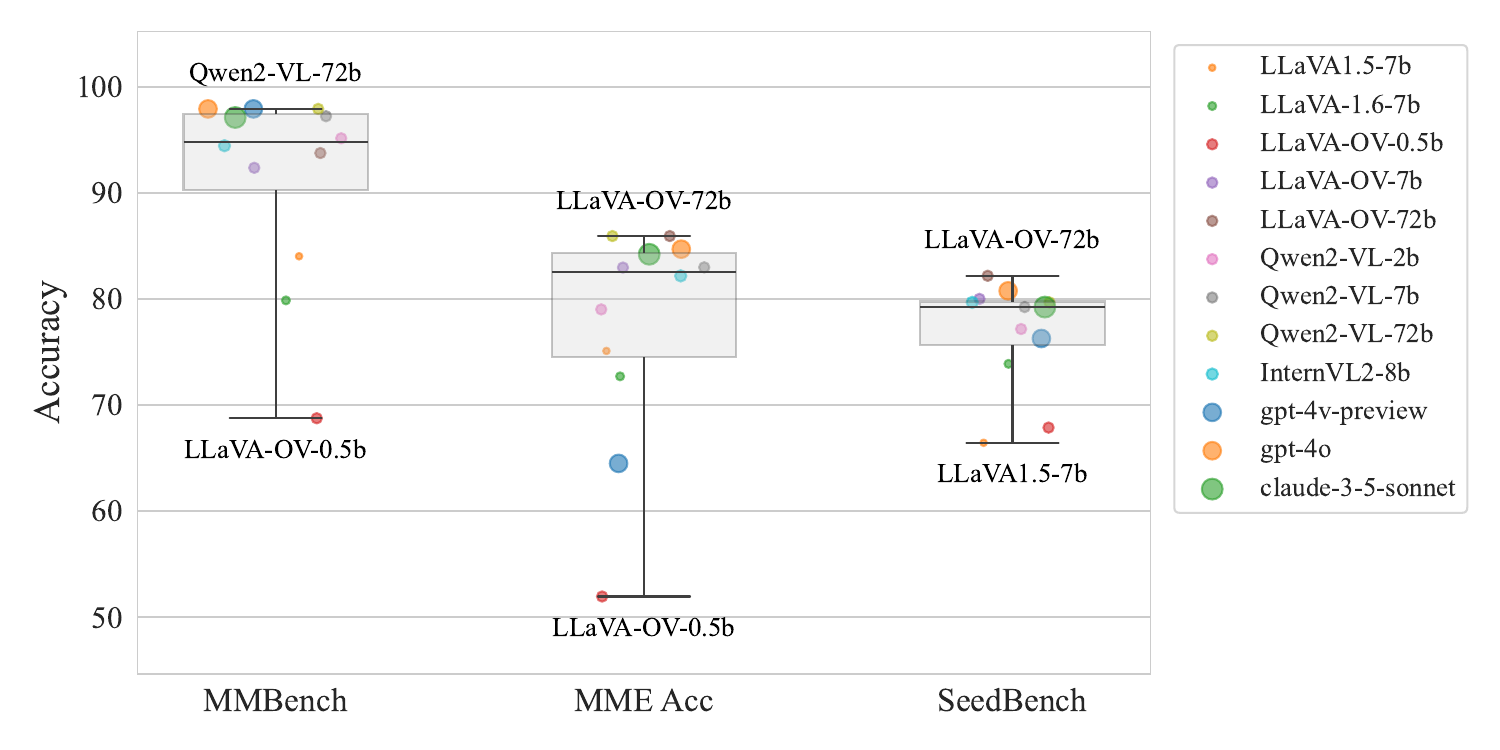} 
        \caption{Evaluation variance across three benchmarks on entity capability.}
        \label{fig:model_performance_analysis}
    \end{subfigure}
    \hfill
    \begin{subfigure}[b]{0.49\textwidth}
        \centering
        \includegraphics[width=0.8\textwidth]{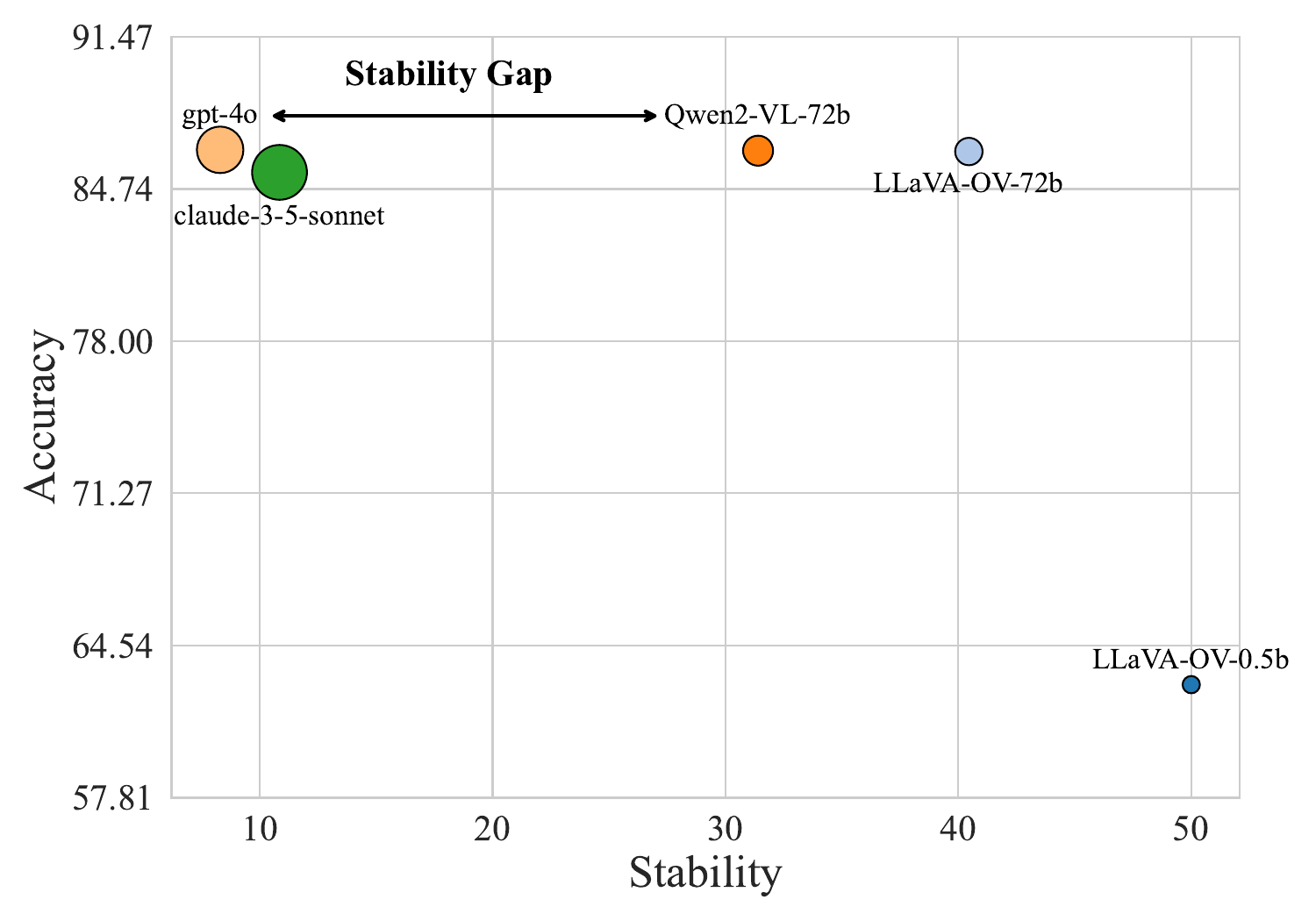} 
        \caption{Comparison on entity capability using our AbilityLens.}
        \label{fig:grounding_performance_vs_stability}
    \end{subfigure}
    
    \caption{\textbf{Motivation behind our AbilityLens and our solution}. (a) shows evaluation variance across three existing benchmarks~\cite{mme, seedbench, MMBench} on entity capability. The differing best and worst models across benchmarks underscore the need for a holistic assessment of perception abilities. (b) Our AbilityLens offers a more complete evaluation by encompassing diverse question types, domains, and metrics, assessing both accuracy and stability of MLLMs. The sizes of closed-source models are artificial estimates.}
     \label{fig motivation}
\end{figure*}

\begin{abstract}

As multimodal large language models (MLLMs) advance rapidly, rigorous evaluation has become essential, providing further guidance for their development. In this work, we focus on a unified and robust evaluation of \textbf{vision perception} abilities, the foundational skill of MLLMs. We find that existing perception benchmarks, each focusing on different question types, domains, and evaluation metrics, introduce significant evaluation variance, complicating comprehensive assessments of perception abilities when relying on any single benchmark. To address this, we introduce \textbf{AbilityLens}, a unified benchmark designed to evaluate MLLMs in six key perception abilities (ranging from counting, OCR, to understanding structural data), focusing on both accuracy and stability, with each ability encompassing diverse types of questions, domains, and metrics.
With the assistance of AbilityLens, we: (1) identify the strengths and weaknesses of current main-stream MLLMs, highlighting stability patterns and revealing a notable performance gap between state-of-the-art open-source and closed-source models; (2) uncover interesting ability conflict and early convergence phenomena during MLLM training; (3) reveal the primary reason of ability conflict is data mixing ratio and LLM model size; and (4) discuss the effectiveness of some straightforward strategies \eg, fine-tuning and model merging, to solve the ability conflict. The benchmark and online leaderboard is released in \url{https://github.com/Chenfeng1271/AbilityLens}.

\end{abstract}

\section{Introduction}
\label{sec:introduction}


The success of large language models (LLM)~\cite{touvron2023llama,bai2023qwen,gpt3.5, gpt4} has inspired strong interest in developing multimodal large language models (MLLMs)~\cite{llava-ov,liu2023llava,dai2024nvlm,qwen2vl}, equipping it with emergent abilities to tackle various vision-language tasks~\cite{docvqa,masry2022chartqa,lu2023mathvista}. 
The key capability that distinguishes MLLMs from LLMs is their ability to `see', \ie, their perception skills.
Thus, evaluating vision perception is particularly critical~\cite{tong2024cambrian}, as it establishes the foundation to understand how effectively MLLMs interpret and comprehend the visual world.

Recent benchmarks~\cite{mme,mmstar,docvqa} evaluate perception capabilities by focusing on varied question types, domains, and metrics, aiming to capture multiple facets of model performance. However, these differences inevitably introduce evaluation variance. As shown in Figure \ref{fig motivation} (a), benchmarks like MME~\cite{mme}, MMBench~\cite{MMBench}, and SeedBench~\cite{seedbench} yield distinct best and worst models, making it challenging to comprehensively assess perception capabilities with any single benchmark. 
Therefore, unlike conventional assessments that overemphasize model accuracy~\cite{ai2d,conbench}, we argue that stability~\cite{atil2024llm,yang2024bounding}, achieving consistent performance across diverse factors such as domains, question types, and metrics, is also important. This metric quantifies the evaluation variance exhibited by models across diverse evaluation scenarios. For example, a robust model excelling in answering True/False questions in MME~\cite{mme} should also handle well multiple-choice questions in SeedBench~\cite{seedbench}. 
Thus, current benchmarks are still insufficient for a comprehensive evaluation of perception capability. 





In this paper, we introduce \textbf{AbilityLens}, a broad-range and efficient benchmark designed to solve these issues. Building on prior perception categories~\cite{mme,wang2024muirbench,mmstar}, AbilityLens assesses six key perception abilities: counting, OCR, attribute recognition, entity extraction, grounding, and structural data understanding.
Our benchmark is constructed through a systematic process. First, to ensure adequate data for each question type, metric, and domain, we compile data from 11 existing benchmarks~\cite{mme,MMBench,seedbench,wang2024muirbench,mmstar,ocrbench,synthdog,masry2022chartqa,ai2d,conbench}, resulting in over 1,000 test samples per ability type and a total of 12,000 test samples overall. Second, we derive sub-metrics from source benchmarks and unify them into a single composite score to measure accuracy and stability. Recognizing that different question types bring distinct baseline performance, \eg, True/False questions have a natural accuracy of 50\% with random choice while four-choice questions only have 25\%, we apply baseline correction to each sub-metric to align performance levels. Then, the model accuracy of each ability is calculated by a weighted sum of these corresponding corrected sub-metrics. Furthermore, we assess the stability of the target model by computing the variance of its z-scores across sub-metrics, which directly reflects its relative performance compared to all candidate models on each sub-metric. To this end, our AbilityLens provides a more comprehensive view on perception capabilities over previous benchmarks, as shown in Figure \ref{fig motivation} (b) and Table \ref{tab difference}.

 Considering our AbilityLens is more general than existing benchmarks, we further use it to monitor the training dynamic by evaluating the intermediate saved training checkpoints, so as to adjust the training strategy to advance MLLMs. Different from current benchmarks~\cite{mme,MMBench}, our AbilityLens provides a more granular understanding of how abilities evolve and interact with each other. Besides, we also compare mainstream MLLMs in each perception ability via absolute ranking. We report 18 state-of-the-art MLLMs~\cite{llava-ov,qwen2vl,internvl2,gpto,claude} on AbilityLens and examine the benefits brought by LLM size, series, and training data. 
We summarize the key experimental findings on AbilityLens: (1) Although current state-of-the-art MLLMs like LLaVA-OV-72b~\cite{llava-ov} and Qwen2VL-72b~\cite{qwen2vl} outperform commercial models~\cite{gpto,claude} in accuracy, there is a notable stability gap between them as shown in Figure \ref{fig motivation} (b), emphasizing the importance of evaluating and improving stability alongside accuracy. (2) We observe the phenomenon of early convergence and potential conflict in abilities during model training.  
Specifically, different perception abilities exhibit different improvement curves, with some abilities meeting performance decline in both accuracy and stability after further training, which is considered an ability conflict. (3) After extensive experiments across various combinations of visual encoders, LLMs, and datasets, we conclude that the data mixing ratio and LLM model size are the primary factors behind the observed ability conflict. Additionally, we conduct a preliminary discussion on leveraging existing fine-tuning and model merging methods to solve the ability conflict.

\begin{table*}[]\caption{\textbf{Comparison between AbilityLens and existing  perception-focused benchmarks.} Sample balance refers to the ratio between the highest and lowest sample counts across different capabilities, while baseline correction assesses performance irrespective of the question type which is necessary for composing multi-source sub-metrics.}
\centering
\resizebox{\linewidth}{!}{
\begin{tabular}{cccccccccccc}
\toprule
 & \multirow{2}{*}{Name} & \multirow{2}{*}{Samples} & \multirow{2}{*}{Sample Balance$\downarrow$} & \multirow{2}{*}{Baseline Correction} & \multicolumn{2}{c}{Eval mode} & \multicolumn{3}{c}{Question Type} & \multicolumn{2}{c}{Performance} \\
 &                       &            &              &                                & Online        & Offline       & T/F      & MCQ      & VQA     & Accuracy    & Stability   \\ \midrule
 & MME~\cite{mme}                   &   2.37k   &   9.9                   &     -                           &   \xmark             &    \cmark           &   \cmark       &     \xmark      &   \xmark       &    \cmark          &      \xmark        \\
 & SeedBench~\cite{seedbench}             &   18k        & 50.8              &     -                          &     \xmark           &   \cmark            &    \xmark        &  \cmark          &   \xmark       &   \cmark            &   \xmark           \\
 & MMBench(en)~\cite{MMBench}               &    11k          &  4.3          &   -                            &   \xmark             &   \cmark            &   \xmark        &  \cmark         &   \xmark       &  \cmark            &      \xmark         \\ 
  & ConBench~\cite{conbench}             &   3k        & 7.1              &      -                          &     \xmark           &   \cmark            &    \xmark        &  \cmark          &   \xmark       &   \cmark            &   \xmark           \\ \midrule
 & UniBench~\cite{al2024unibench}              &   -          &     -        &     \xmark                           &      \cmark          &  \cmark             &    \cmark      &    \cmark      &    \cmark     &      \cmark       &      \xmark         \\
 & EUREKA~\cite{balachandran2024eureka}                &   83k           &   38         &      \xmark                           &    \cmark           &   \cmark            &  \cmark         &   \cmark        &   \cmark       &    \cmark          &       \xmark         \\
\rowcolor{blue!15}  & AbilityLens(ours)     &    12k       &     2.6          &                \cmark                &    \cmark           &     \cmark          &     \cmark     &  \cmark        &      \cmark   &   \cmark          &   \cmark      \\ \bottomrule   
\end{tabular}}\label{tab difference}
\end{table*}

\section{Related Work}

\noindent \textbf{Perception-based MLLM benchmarks.} In parallel with architectural advancements~\cite{llava-ov,Qwen2.5-VL}, considerable effort has been dedicated to improving benchmarks for MLLMs, which play a crucial role in steering the development of the next generation of models~\cite{mmmu}. As a fundamental capability, perception is essential to the performance of MLLMs. To evaluate this skill, benchmarks such as MME~\cite{mme}, MMStar~\cite{mmstar}, MMBench~\cite{MMBench}, and SeedBench~\cite{seedbench} have been introduced, aiming to capture data from diverse domains and cover a wide range of question types. However, these differences inevitably introduce evaluation variance, motivating the development of a unified dataset that enables a comprehensive assessment of perception abilities.

\noindent \textbf{Ability-perspective MLLM benchmarks.} 
To enable a systematic evaluation of MLLM advancements, recent benchmarks~\cite{moayeri2024unearthing,al2024unibench} aim to expose granular weaknesses within specific capabilities by regrouping benchmarks~\cite{mvbench,docvqa,mmmu} according to model abilities. For instance, Unibench~\cite{al2024unibench} categorizes 53 vision-language benchmarks by different capabilities, though most focus on traditional recognition tasks and are limited in evaluating state-of-the-art MLLMs. EUREKA~\cite{balachandran2024eureka}, on the other hand, concentrates on non-saturated samples within each ability, offering a more targeted view of areas most ripe for improvement. However, these current benchmarks primarily emphasize offline model comparisons, providing limited insight into when and why models may underperform on specific capabilities or efficient ways to enhance these weaker perception capabilities.


\noindent \textbf{Multimodal large language models.} Building upon the success of LLMs such as GPT~\cite{radford2018improving}, LLaMA~\cite{touvron2023llama}, and Qwen~\cite{bai2023qwen}, and vision foundation models including DINOv2~\cite{oquab2023dinov2}, CLIP~\cite{clip}, and SigLIP~\cite{siglip}, MLLMs~\cite{llava-ov,liu2023llava,liu2024points,lu2024ovis} have emerged as powerful tools for multimodal understanding and reasoning. However, training a versatile MLLM presents significant challenges~\cite{jiang2024mantis}. Current state-of-the-art approaches~\cite{liu2023improvedllava,internvl2} primarily rely on more powerful LLMs and vision encoders coupled with extensive training data, requiring thousands of GPU hours for each training run. While these resource-intensive solutions have yielded improvements, we still lack a clear understanding of how these models evolve during training and how different components interact with each other. This opacity in model behavior impedes us to develop more effective strategies for advancing MLLM capabilities.

\section{AbilityLens}
\label{sec:methodology}

\begin{table*}[!h]
\resizebox{\linewidth}{!}{
\begin{tabular}{c|cccccc}
\toprule
Ability                    & Source    & Subcategory              & Question Type & Metric                   & Sample                & Total Sample \\ 
\midrule
\multirow{5}{*}{Counting}  & MME       & counting                    & T/F  & mme\_score               & 60                    & \multirow{5}{*}{1759} \\
                           & SeedBench & instances counting       & MCQ  & seed\_img               & 2447$\rightarrow$1223                  &                  \\
                           & MMStar    & object counting          & MCQ  & fine\_grained\_perception  & 92                    &                  \\
                           & Muirbench & counting                 & MCQ  & muirbench\_score         & 234                   &                  \\
                           & Mirb      & counting                 & VQA  & mirb\_score              & 150                   &                  \\ 
\midrule
\multirow{4}{*}{OCR}       & MME       & ocr                      & T/F  &    mme\_score                     & 40                    & \multirow{4}{*}{1844} \\
                           & MMBench   & ocr                      & MCQ  &   gpt\_eval\_score                      & 304                   &                  \\
                           & OCRBench  & ocr                      & VQA  &   accuracy                      & 1000                  &                  \\
                           & Synthdog  & ocr                      & VQA  &   tree\_edit\_distance                      & 500                   &                  \\ 
\midrule
\multirow{5}{*}{Grounding} & MME       & position                 & T/F  &         mme\_score                & 60                    & \multirow{5}{*}{1493} \\
                           & SeedBench & position, spatial relation & MCQ &  seed\_img                       &      1636 $\rightarrow$ 817                &                  \\
                           & MMStar    & localization             & MCQ  &   fine\_grained\_perception                      & 40                    &                  \\
                           & MMBench   & position, spatial relation & MCQ &     gpt\_eval\_score                    &       492                &                  \\
                           & Muirbench & localization             & MCQ  &   muirbench\_score                      & 84                    &                  \\ 
\midrule
\multirow{3}{*}{Entity}    & MME       & landmark, celebrity      & T/F  &            mme\_score             & 740                   & \multirow{3}{*}{2227} \\
                           & SeedBench & identity reasoning       & MCQ  &    seed\_img                     &    1831 $\rightarrow$   915                &                  \\
                           & MMBench   & landmark                 & MCQ  &     gpt\_eval\_score                    &         572              &                  \\ 
\midrule
\multirow{4}{*}{Attribute} & MMStar    & attribute reasoning      & MCQ  &     fine\_grained\_perception                    & 188                   & \multirow{4}{*}{3453} \\
                           & Muirbench & difference spotting      & MCQ  &      muirbench\_score                   & 536                   &                  \\
                           & MMBench   & attribute recognition and comparison & MCQ &gpt\_eval\_score &    405                   &                  \\
                           & SeedBench & instance attributes      & MCQ  &     seed\_img                    &      4649 $\rightarrow$ 2324                &                  \\ 
\midrule
\multirow{5}{*}{Structured Data} 
                           & ChartQA   & chart                    & VQA  &   relaxed\_overall                      & 2500 $\rightarrow$ 400 & \multirow{5}{*}{1332} \\
                           & AI2D      & graph                    & MCQ  &   exact\_match                      & 3088 $\rightarrow$ 300 &                  \\
                           & ConBench  & poster                   & VQA  &  ConScore\_D                       & 240                   &                  \\
                           & MMBench   & code, map, diagram       & MCQ  &      gpt\_eval\_score                   & 282                   &                  \\
                           & MMStar    & diagram                  & MCQ  & fine\_grained\_perception                        & 110                   &                  \\ 
\bottomrule          
\end{tabular}}
\caption{\textbf{The data component detail of AblilityLens.} Data of AbilityLens is sourced from 11 public benchmarks. We reduce the sample size in large datasets, where $\rightarrow$ indicates the number of samples before and after reduction. We also provide further decomposition in the supplementary material.}\label{tab overview}
\label{tab:overview}
\end{table*}

In this section, we elaborate on the details of our AbilityLens. Our AbilityLens includes (1) dataset collection and metrics, and (2) evaluation framework with online and offline modes.

\subsection{Dataset Collection and Metrics}

\noindent \textbf{Dataset collection.} A major challenge lying in building a comprehensive ability-focused evaluation is the lack of a unified evaluation protocol, as model and benchmark creators often implement custom pipelines that vary widely in data preparation, output postprocessing, and metric calculation. Therefore, we develop our AbilityLens based on LMMs-Eval~\cite{zhang2024lmms}, a comprehensive evaluation suite for VLMs that supports over 90 VLM benchmarks. In detail, as illustrated in Table \ref{tab overview}, AbilityLens focuses on 6 core perception abilities following prior works~\cite{mme,MMBench}: counting, OCR, attribute recognition, entity extraction, grounding, and structural data understanding. We select relevant 11 datasets from LMMs-Eval to construct a uniform one for each ability. The included benchmarks encompass MME~\cite{mme}, MMBench~\cite{MMBench}, SeedBench~\cite{seedbench}, MuirBench~\cite{wang2024muirbench}, Mirb~\cite{mirb}, MMStar~\cite{mmstar}, OCRBench~\cite{ocrbench}, SynthDog~\cite{synthdog}, ChartQA~\cite{masry2022chartqa}, AI2D~\cite{ai2d}, and ConBench~\cite{conbench}, covering diverse domain, question types and metrics. To achieve efficient yet broad-ranging evaluation, we follow LMMs-Eval~\cite{zhang2024lmms} to remove the redundant samples, allowing users to obtain the results of a 7B MLLM in 0.5 hour with 8 H100 GPUs.

\noindent \textbf{Evaluation metric.} For each ability, we define two performance metrics—model accuracy and model stability—each computed from sub-metrics extracted from the source benchmarks. Because different question types have varying baseline performance levels, representing the expected accuracy from random guessing, we first adjust each sub-metric to account for these differences. For instance, True/False questions have a baseline accuracy of 50\% due to the binary choice, while multiple-choice questions may have a lower baseline depending on the number of options. To ensure that the true ability score remains independent of question type, we apply baseline correction to each sub-metric, aligning them to a consistent standard. Let \( m_{ij} \) represent the result of model $i$ on the \( j \)-th sub-metric of an ability, with \(\text{UB}_{j}\) denoting the upper bound of the metric and \(\text{BL}_{j}\) as its baseline performance\footnote{The baseline performance for the VQA benchmark is set to 0.}. The baseline correction for the result \( m_{ij} \) is then defined as:

\begin{equation}
\tilde{m}_{ij}  = \frac{m_{ij} - \text{BL}_{j}}{\text{UB}_{j} - \text{BL}_{j}}.
\end{equation}

The accuracy of model $i$ for each ability, denoted as $\mathcal{A}_i$, is then calculated as a weighted sum of corresponding corrected sub-metrics $\tilde{m}_{ij} $, weighted by their sample counts:

\begin{equation}
\mathcal{A}_i = \sum_{j} \frac{n_j \cdot \tilde{m}_{ij} }{N},
\end{equation}
where $N=\sum_{j}{n_j}$ is the total sample count and $n_j$ is the sample count for the $j$-th metric. To assess the model stability of each ability, we first use z-score $z_{ij}$ to measure the relative performance of model $i$ with other model candidates in each sub-metric:
\begin{equation}
    z_{ij} = \frac{m_{ij}-\mu_j}{\sigma_j},
\end{equation}

where $\mu_j$ and $\sigma_j$ are the mean and variance of the results of model candidates on this sub-metric, respectively. In this paper, we use 18 MLLMs from different series, sizes, sources, and LLMs as the candidates, capturing a wider spectrum of architectural variations and performance characteristics. Then, we measure the standard deviation of the resulting z-scores as the stability score $\mathcal{I}$ of an ability,  with higher values indicating greater instability:
\begin{equation}
    \mathcal{I}_i =\operatorname{std}(\mathbf{z}_{i}). 
\end{equation}

\subsection{Evaluation Framework}
\begin{figure*}
    \centering
    \includegraphics[width=1\linewidth]{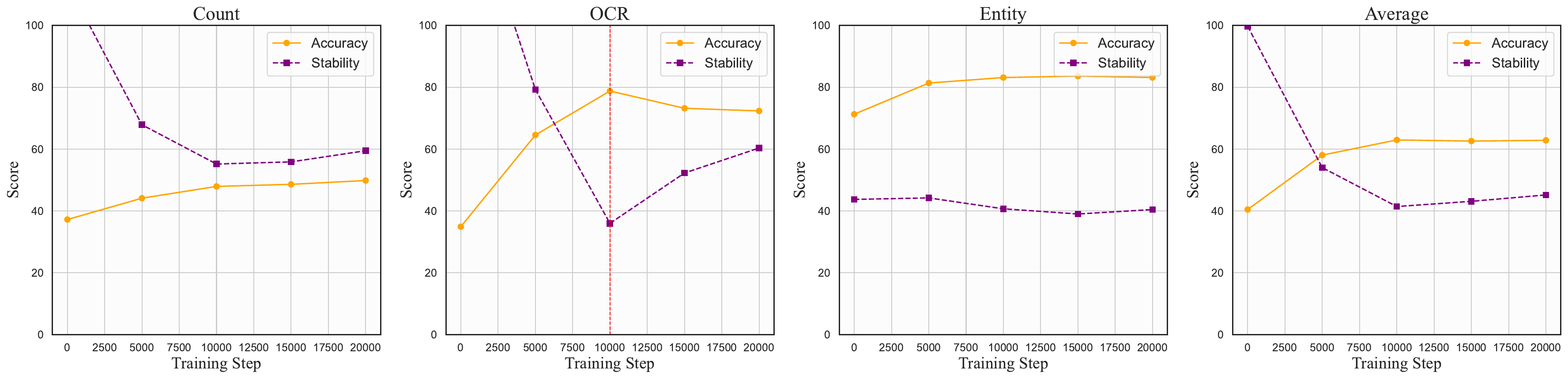}
    \caption{\textbf{Monitoring training dynamics with AbilityLens.} Unlike traditional methods that focus solely on average scores, often obscuring underlying conflicts, AbilityLens tracks optimization trends for each specific ability, revealing an ability conflict indicated by a significant performance degradation in accuracy and stability of OCR performance after 10,000 steps.}
    \label{fig dynamic}
\end{figure*}

Our AbilityLens offers both online and offline evaluation modes to accommodate diverse model assessment needs.

\noindent \textbf{Offline Evaluation Mode} provides a comprehensive assessment of each model's perception capabilities in accuracy and stability, as shown in Table \ref{tab accuracy} and \ref{tab: stability}, respectively. This approach effectively reduces evaluation bias and variance caused by the single benchmark, enabling a more robust and holistic comparison across models. We evaluate state-of-the-art MLLMs with various model sizes and training data, including both commercial and open-source models, across six core abilities.  

\noindent \textbf{Online Evaluation Mode} enables real-time monitoring of training dynamics throughout model development, leveraging AbilityLens for a more unified and comprehensive assessment compared to existing benchmarks. By evaluating checkpoints at various training stages, we can (1) assess the impact of training configuration adjustments, (2)
identify optimal checkpoints for specific abilities, (3)
track trends in ability-specific optimization, and (4)
detect potential \textbf{ability conflicts where some abilities meet performance degradation in both accuracy and stability after further training}. Addressing these conflicts during training can lead to more balanced and stable multimodal capabilities, ultimately enhancing the overall effectiveness of the models.
In Figure \ref{fig dynamic}, we illustrate AbilityLens in action through a reimplementation of LLaVA-OV-SI’s~\cite{llava-ov} single-stage training process. Traditional approaches, represented by the fourth subfigure in Figure \ref{fig dynamic}, focus solely on average scores and often obscure underlying conflicts. In contrast, AbilityLens offers a capability-decomposed analysis tool that enables researchers to gain deeper insights and make targeted improvements in their VLMs. For example, in the second figure of Figure \ref{fig dynamic}, after 10,000 training steps, the OCR ability displays a marked performance degradation in both accuracy and stability, revealing an ability conflict that conventional methods might overlook. 
We will discuss using existing fine-tuning and model merging methods to solve this issue in Sec \ref{sec discussion}. 

\begin{table*}[!t]\caption{\textbf{Accuracy comparison.} Accuracy $\uparrow$ of 18 MLLMs on AbilityLens where cells highlighted in \cellcolor{green!20}{green} indicate the best performance, and \cellcolor{red!20}{red} indicate the worst performance.}
\centering
\resizebox{\linewidth}{!}{
\begin{tabular}{l|ccccccc}
\toprule
Model                        & Counting   & OCR         & Grounding   & Entity      & Attribute   & Structured Data & Average \\ \midrule
LLaVA1.5-7b                 & 36.78      & \cellcolor{red!20}29.13       & 28.64       & 73.83       & 46.88       & \cellcolor{red!20}22.87           & \cellcolor{red!20}39.69 \\
LLaVA1.6-7b                  & 42.80      & 63.99       & 38.80       & 75.02       & 55.70       & 37.89           & 52.37 \\
LLaVA-OV-0.5b                & \cellcolor{red!20}32.33      & 64.54       & \cellcolor{red!20}17.04       & \cellcolor{red!20}62.81       & \cellcolor{red!20}44.90       & 23.55           & 40.86 \\
LLaVA-OV-7b                 & 51.60      & 71.23       & 50.36       & 84.16       & 62.04       & 64.83           & 64.04 \\
LLaVA-OV-SI-7b               & 49.44      & 78.17       & 45.24       & 85.55       & 62.22       & 61.40           & 63.67 \\
LLaVA-Video-7b               & 37.82      & 53.08       & 32.00       & 68.56       & 60.63       & 33.13           & 47.54 \\
LLaVA-OV-72b                 & \cellcolor{green!20}56.75 & 81.35       & 59.23 & 86.40       & \cellcolor{green!20}69.08 & 73.15           & 70.99 \\
Qwen2VL-2b                 & 48.28      & 73.36       & 35.00       & 82.39       & 56.61       & 46.36           & 57.00 \\
Qwen2VL-7b                   & 50.95      & 79.29       & 47.12       & 85.10       & 62.66       & 67.56           & 65.45 \\
Qwen2VL-72b                  & 55.84      & \cellcolor{green!20}86.09       & 58.70       & 86.43       & 65.11       & \cellcolor{green!20}79.32           & 71.92 \\
Qwen2.5VL-3b                 &  50.27     & 77.49       &    40.01     &    82.56    &   60.70     &   56.10         &  61.19  \\
Qwen2.5VL-7b                   &  50.52     &  80.10      &   47.71      &   86.19     &   62.60     &  72.44          &   66.59   \\
Qwen2.5VL-72b                  &   56.53    &  85.70       &  \cellcolor{green!20} 59.62     & \cellcolor{green!20} 86.82      &   67.11     & 79.20           & \cellcolor{green!20} 72.49  \\
InternVL2-8b               & 49.23      & 80.47       & 50.89       & 84.30       & 61.21       & 64.92           & 65.17 \\
InternVL3-8b               &  51.12      &  80.02       &   51.45      &    84.55     &  63.46       &   70.12          &  66.79  \\
gpt-4-vision-preview        & 39.00      & 71.95       & 47.67       & 77.92       & 58.56       & 67.12           & 60.37 \\
gpt-4o-2024-08-06            & 50.55      & 82.57       & 58.84       & 86.47       & 67.67       & 75.82           & 70.32 \\
claude-3-5-sonnet-2024102    & 51.64      & 79.80       & 58.63       & 85.47       & 61.20       & 77.06           & 68.97 \\ \bottomrule
\end{tabular}}
\label{tab accuracy}
\end{table*}

\begin{table*}[!t]
\caption{\textbf{Stability comparison.} Stability score $\downarrow$ of 18 MLLMs on AbilityLens, with cells highlighted in \cellcolor{red!20}red for the highest and \cellcolor{green!20}green for the lowest scores. Lower scores indicate better stability.}
\centering
\resizebox{\linewidth}{!}{
\begin{tabular}{l|ccccccc}
\toprule
Model                        & Counting   & OCR         & Grounding   & Entity      & Attribute   & Structured Data & Average \\ \midrule
LLaVA1.5-7b                  &  72.54      &\cellcolor{red!20} 128.40       & 42.32       & 110.40        & \cellcolor{red!20} 85.30        & 46.73           & 80.95 \\
LLaVA1.6-7b                  & 48.00      & 72.73       & \cellcolor{green!20} 31.47       & 57.29       & 28.37       & 43.95           & 46.96 \\
LLaVA-OV-0.5b                &  67.07      & 110.5       & 60.61       & 50.00          & 82.58       & 73.21           & 74.00 \\
LLaVA-OV-7b                  &  65.09      & 62.05       & 49.28       & 26.81       & 72.56       & \cellcolor{green!20} 18.62           & 59.64 \\
LLaVA-OV-SI-7b               & 115.80      & 34.17       & 86.64       & 42.46       & 53.20        & 25.74           &  \cellcolor{red!20} 88.54 \\
LLaVA-Video-7b               & \cellcolor{red!20} 152.20      & 60.16       & 78.28       & \cellcolor{red!20}  132.30        & 28.56       & \cellcolor{red!20} 79.74           &  56.43 \\
LLaVA-OV-72b                 & 96.88      & 57.49       & 58.85       & 40.45       & 42.29       & 42.63           & 56.43 \\
Qwen2VL-2b                   & 47.50      & 89.28       & 56.53       & 20.97       & 64.92       & 46.80            & 54.33 \\
Qwen2VL-7b                   & \cellcolor{green!20} 44.08      & 45.69       & 57.59       & 12.01       & 46.24       & 27.55           & 38.86 \\
Qwen2VL-72b                  & 88.18      & 24.78       & 51.52       & 31.40        & 49.34       & 27.82           & 44.56 \\
Qwen2.5VL-3b                 &   45.16    &   72.17     &  60.24       &  18.88      &   52.40     &   47.09         & 49.32  \\
Qwen2.5VL-7b                   &  45.10     &    46.70    &     56.91    &   14.09     &    47.28    &  22.17          &   38.71   \\
Qwen2.5VL-72b                  &   82.81    & 29.70      &     49.62   &    39.89    &  43.10      & 28.33          &  45.57  \\
InternVL2-8b                 & 44.16      & 17.34       & 47.93       & 9.74        & 54.60        & 60.48           & 38.54 \\
InternVL3-8b               &   49.12     &    24.63     &   44.41      &     13.82    &  51.17       &  62.05           &   40.87 \\
gpt-4-vision-preview         & 85.72      &  77.08       &\cellcolor{red!20} 93.53       & 80.34       & 52.74       & 55.78           & 64.20 \\
gpt-4o-2024-08-06            &  86.98       & \cellcolor{green!20} 24.01       & 57.06       & \cellcolor{green!20} 8.30         & \cellcolor{green!20} 12.29       & 36.48           & \cellcolor{green!20} 37.52 \\
claude-3-5-sonnet-2024102    &  84.57      & 24.78        & 59.68       & 10.85       & 25.86       & 27.80            & 38.92 \\ \bottomrule
\end{tabular}}
\label{tab: stability}
\end{table*}

\section{Experiments}

\begin{figure*}[t]
    \centering
    \includegraphics[width=1\linewidth]{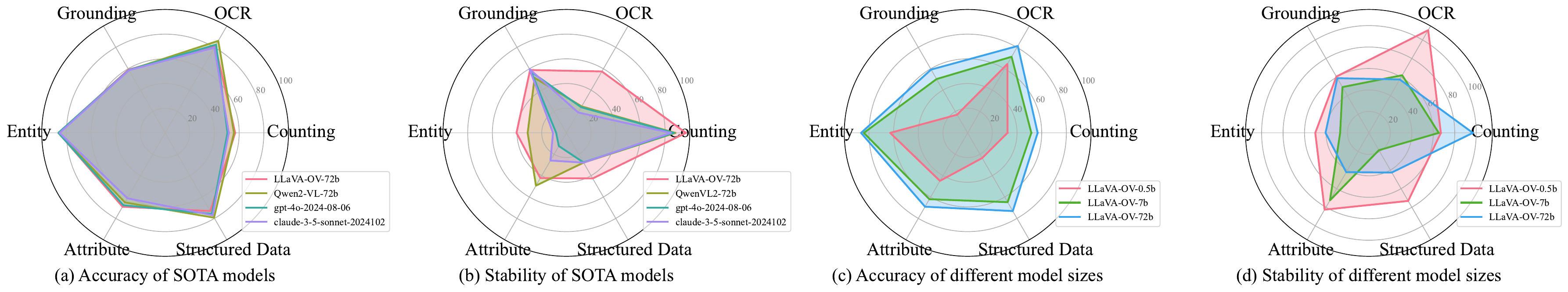}
    \caption{\textbf{Accuracy and stability comparison} across AbilityLens's six perception dimensions, evaluating four state-of-the-art MLLMs alongside models of different LLM scales.}
    \label{fig sota}
\end{figure*}

\noindent \textbf{Implementation Details.} 
For offline evaluation, we evaluate 18 MLLMs with various model sizes and training data, including three strong commercial models: gpt-4-vision~\cite{gpt4v}, gpt-4o~\cite{gpto}, and claude-3-5-sonnet~\cite{claude}, as well as 15 open-source state-of-the-art models, including LLaVA 1.5~\cite{liu2023improvedllava}, LLaVA 1.6~\cite{liu2024llavanext}, the LLaVA-OV series~\cite{llava-ov}, the Qwen2VL series~\cite{qwen2vl}, Qwen2.5VL series~\cite{Qwen2.5-VL} and the InternVL series~\cite{internvl2}. 
The detailed model versions and evaluation results are presented in Table \ref{tab accuracy} and \ref{tab: stability}, and further discussed in Sec. \ref{sec offline exp}. 
For online evaluation, we implement LLaVA-OV-SI with Qwen2-7b~\cite{bai2023qwen} as the LLM backbone, using the official codebase~\cite{llava-ov} on 128 H100 GPUs. For continuous fine-tuning of LLaVA-OV-SI-Qwen2-7b to enhance weaker capabilities, we retain the original training configuration and initialize the model from the final 20,000-step checkpoint. To mitigate catastrophic forgetting, we adjust the sampling ratio by setting general data to 10\% and increasing OCR data to 20\% for the single-image dataset of LLaVA-OV. For using model merging method Task  Arithmetic~\cite{ilharco2022editing} to alleviate ability conflict, we first collect a small calibration set with 3000 OCR samples from the single-image dataset of LLaVA-OV and tune the 1,0000-step checkpoint on it to find the task arithmetic. Then we add the arithmetic to the final checkpoint. 

\begin{figure*}[ht]
    \centering
    \includegraphics[width=1\linewidth]{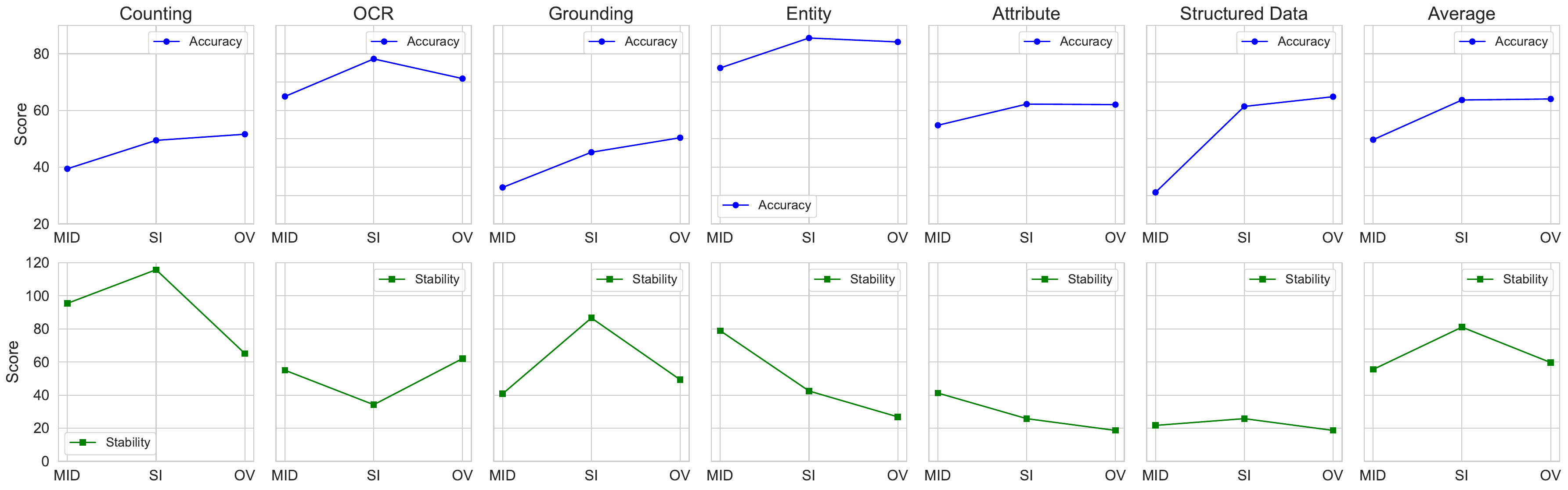}
    \caption{\textbf{Performance dynamics in the middle (MID), single-image (SI), and one-vision (OV) stages 
    of LLaVA-OV-7b.} Accuracy curves demonstrate consistent improvement across stages, except for OCR tasks, whereas stability curves indicate that stability is less prioritized in data preparation.}
    \label{fig:stage data}
\end{figure*}
\subsection{Offline Evaluation}\label{sec offline exp}
\noindent \textbf{Overall accuracy.} Table \ref{tab accuracy} presents a detailed breakdown of 18 MLLMs, 
varying in model size, training data, and LLM backbone, evaluated on our AbilityLens. We highlight the following key findings regarding accuracy: (1) No single model demonstrates superior accuracy across all abilities, suggesting that achieving a model that excels in all abilities is non-trivial.
(2) These 6 perception tasks vary in difficulty for MLLMs. For example, the best performances on OCR and entity tasks exceed 80\%, while for counting and grounding tasks, they are below 60\%. It suggests that curriculum learning could be employed to strategically reorder the training sequence, beginning with simpler tasks and progressively advancing to more challenging ones.

\noindent \textbf{Overall stability.}
Table \ref{tab: stability} and \ref{fig sota} (b)
report the stability comparison of SOTA models, highlighting a different ranking compared to accuracy. We find that closed-source models generally achieve lower stability scores than open-source models, even attaining similar accuracy levels. For example, Qwen2-VL-72b~\cite{qwen2vl} has an average accuracy score of 71.92 and an average stability score of 44.56, while gpt-4o-2024-08-06~\cite{gpto} has an accuracy of 70.32 and a stability score of 37.52. These results suggest that closed-source models are more stable than open-source models, a factor often neglected in accuracy-only benchmarks. This underscores the necessity of stability measurement to prevent model design from overfitting single performance rankings. \\

\vspace{-2mm}

\subsection{Online Evaluation}
In this section, we first train an LLaVA-OV-SI model from scratch and use online evaluation to track any changes in both accuracy and stability.\\
\noindent \textbf{Accuracy and stability tracking.} 
 Figure \ref{fig dynamic} illustrates (1) the learning progression of perception abilities differs from each other, with model skills converging at different stages of training. For instance, entity recognition stabilizes in the early steps, OCR performance peaks mid-training, and counting accuracy is highest in the later steps. (2) Stability also varies across skills. For example, in counting and OCR, instability even increases as training progresses.

\section{Discussion and Analysis}\label{sec discussion}

\noindent \textbf{Reason behind ability conflict.} To investigate the underlying causes of ability conflict, we evaluate various combinations of visual encoders, LLMs, and datasets. Detailed results are provided in Figure \ref{fig trend}, \ref{fig data} below, and Figure 3 of the supplementary material. Notably, \textbf{the reasons for ability conflict are model size and data mixing}, where different visual encoder+LLM combinations exhibit similar performance trends on the same dataset, while variations in the data mixing ratio and different LLM sizes significantly impact performance changes and even alter the timing and number of ability conflict. Specifically, a weaker LLM can result in more ability conflict in training.

\noindent \textbf{Analyzing the effect of model size.}
To investigate the effect of the model size, we report LLaVA-OV accuracy and stability score with 3 different sizes in Figure \ref{fig sota} (c) and (d), respectively. Although it is generally accepted that a larger language model base enhances performance, we find that this relationship holds primarily for accuracy, not stability across certain abilities where stability in grounding and counting shows an opposite trend.

\begin{figure*}[t]
    \centering
    \begin{subfigure}[b]{\textwidth} 
        \centering
        \includegraphics[width=1.\textwidth]{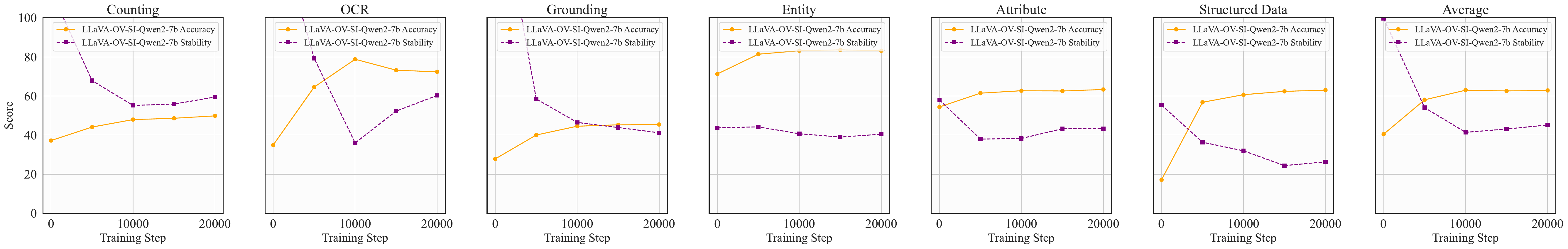}
        \label{fig:subfig1}
    \end{subfigure}

    \begin{subfigure}[b]{\textwidth}
        \centering
        \includegraphics[width=1.\textwidth]{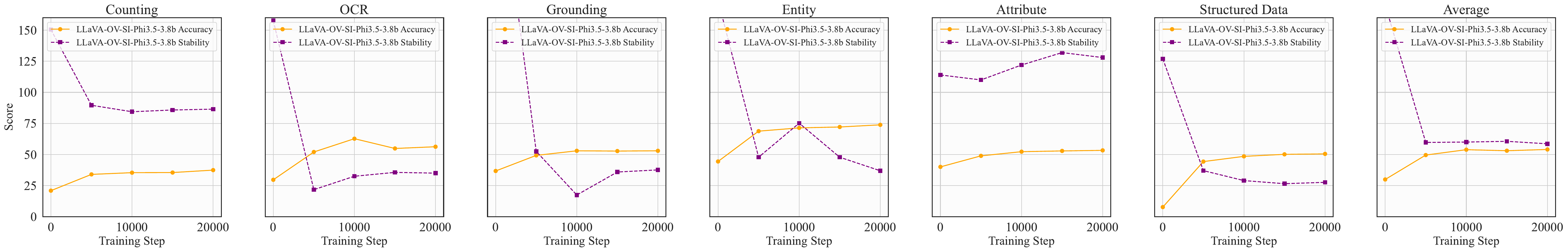}
        \label{fig:subfig2}
    \end{subfigure}

    \begin{subfigure}[b]{\textwidth}
        \centering
        \includegraphics[width=1\textwidth]{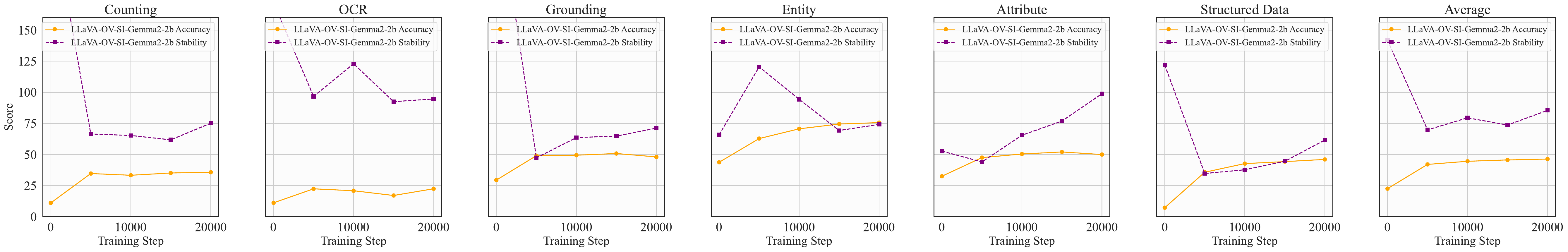}
        \label{fig:subfig3}
    \end{subfigure}

    \caption{Monitoring the training dynamics of LLaVA-OV-SI using Qwen2-7b (first row), Phi3.5-3.8b (second row), and Gemma2-2b (third row) across six abilities. We find that model size significantly impacts the occurrence of ability conflict.}
    \label{fig trend}
\end{figure*}

\noindent \textbf{Analyzing the effect of model version and visual encoder.} We further investigate the influence of visual encoder and LLM in Table \ref{decompose} and observe an interesting phenomenon: (1) A strong LLM (e.g., Qwen2 vs. Phi3.5) primarily enhances accuracy, but may not be optimal for ensuring the stability of certain abilities. (2) A strong visual encoder (CLIP vs. SigLIP) can boost both accuracy and stability performance.

\begin{table*}[h]
\centering
\caption{Ablation study on visual encoder and LLM.}\label{decompose}
\resizebox{0.7\linewidth}{!}{
\begin{tabular}{lccccccc}
\toprule
\multicolumn{8}{c}{\textbf{Accuracy}} \\ \midrule
Models & Counting & OCR & Grounding & Entity & Attribute & Structured & Average \\ \midrule 
 CLIP-L + Qwen2-7B       &  46.32 & 69.91 & 44.29 & 78.30 & 60.19 & 62.22 & 60.21 \\
 Siglip-400M + Qwen2-7B   &  49.87 & 72.37 & 45.44 & 83.18 & 63.35 & 63.01 & 62.87 \\ 
 Siglip-400M + Phi3.5-3.8B  &  37.42 & 56.23 & 52.97 & 73.86 & 53.31 & 50.47 & 54.05 \\ \midrule 
\multicolumn{8}{c}{\textbf{Stability}} \\ \midrule
 CLIP-L + Qwen2-7B       &  60.21 & 65.35 &  38.80 &  48.98 & 47.35 & 26.70 & 47.90  \\
 Siglip-400M + Qwen2-7B   &  59.5 & 60.38 &  41.20 &  40.48 & 43.30 & 26.35 &  45.20 \\ 
 Siglip-400M + LPhi3.5-3.8B   &  86.49 & 35.04 & 37.60 &  36.91 & 128.45 & 27.61 & 58.64 \\  
\bottomrule
\end{tabular}}
\end{table*}

\noindent \textbf{Investigating the effect of training data.}
Existing MLLM training typically adopts a multi-stage strategy, utilizing different types of training data at each stage. Using AbilityLens, we analyze the impact of each stage by reporting the accuracy and stability scores for various abilities in LLaVA-OV, as shown in Figure \ref{fig:stage data}. Our findings reveal that current methods for developing training datasets are heavily focused on improving accuracy, while stability remains inconsistent and less prioritized.

\noindent \textbf{Simple solutions to alleviate ability conflict.} When conflict arises in only one ability during training, approaches such as continuous fine-tuning or model merging can help mitigate the issue. As demonstrated in Tables \ref{tab: online acc} and \ref{tab: online stab}, Task Arithmetic~\cite{ilharco2022editing} generally improves both accuracy and stability compared to continuous fine-tuning, which may be sensitive to data ratio variations. However, addressing conflicts across multiple abilities remains an open challenge, as multi-model merging struggles to preserve the optimal performance of each individual model.

\begin{figure*}[tbp]
    \centering
    \begin{subfigure}[b]{\textwidth} 
        \centering
        \includegraphics[width=1.\textwidth]{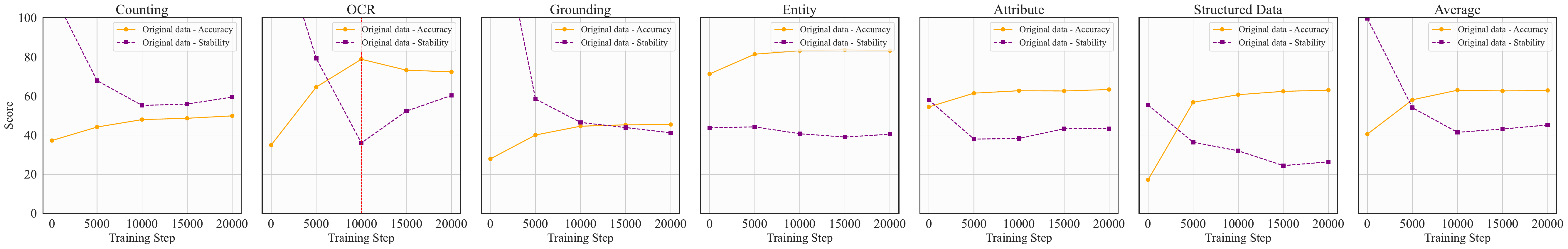}
        \label{fig:subfig1}
    \end{subfigure}

    \begin{subfigure}[b]{\textwidth}
        \centering
        \includegraphics[width=1.0\textwidth]{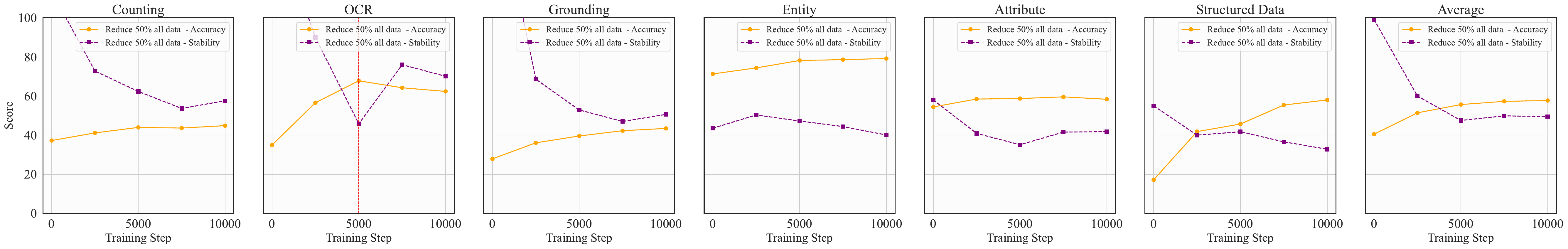}
        \label{fig:subfig2}
    \end{subfigure}

    \begin{subfigure}[b]{\textwidth}
        \centering
        \includegraphics[width=1.0\textwidth]{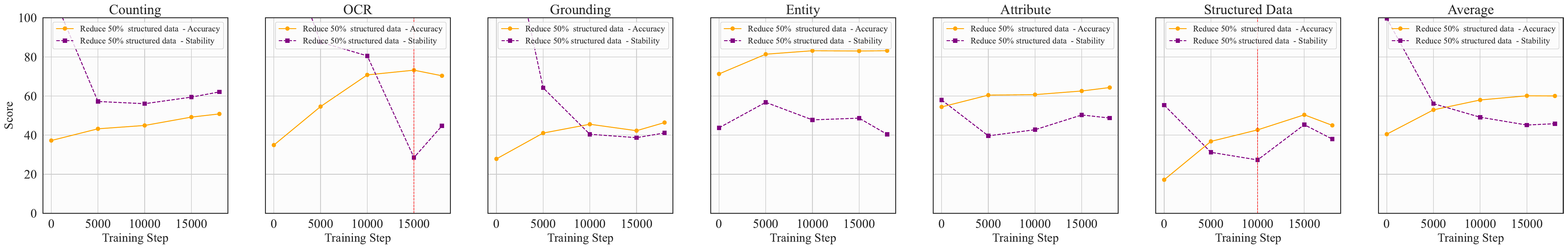}
        \label{fig:subfig3}
    \end{subfigure}

    \caption{Monitoring training dynamics of LLaVA-OV-SI using original data (first row), 50\% original data (second row), and original data reducing 50\% structured data (third row) across six abilities. }
    \label{fig data}
\end{figure*}

\begin{table*}[t!]\caption{Accuracy comparison $\uparrow$ with continuous fine-tuning and model merging methods in improving the weak OCR ability caused by ability conflict. Notably, the final step checkpoint is deemed as the baseline.}
\centering
\resizebox{\linewidth}{!}{
\begin{tabular}{ccccccccc}
\toprule
       & GPU hours & Counting & OCR & Grounding & Entity & Attribute & Structured & Average \\ \midrule
 LLaVA-OV-SI-Qwen2-7b (10000 step) &  -      &      47.96               &     78.82         &    44.58                   &   83.18                &     62.73                 &        60.70               & 62.99 \\
 LLaVA-OV-SI-Qwen2-7b (final step)  & -      &  49.87                     &  72.38                & 45.44                      &  83.18                   &  63.36                      &  63.01                    & 62.87 \\  \midrule
Continuous Fine Tuning &  665.6      &    47.10                  & 81.77                &     44.79                  &   81.22                 &   63.25                     &    62.34                & 63.41 \\  
Task Arithmetic(addition)~\cite{ilharco2022editing} &     13.7   &    49.60                  &    78.19              &   45.50                    &      83.46               &    63.50                     &     63.17              &  63.90  \\    
\bottomrule
\end{tabular}}
\label{tab: online acc}
\end{table*}

\begin{table*}[t!]\caption{Stability comparison $\downarrow$ with continuous fine-tuning and model merging nethods in improving the weak OCR ability caused by the ability conflict.}
\centering
\resizebox{\linewidth}{!}{
\begin{tabular}{ccccccccc}
\toprule
       & GPU hours & Counting & OCR & Grounding & Entity & Attribute & Structured & Average \\ \midrule
 LLaVA-OV-SI-Qwen2-7b (10000 step) &  -      &  55.20              &  36.01         &  46.50           &  40.72           &  38.30              &  31.99           &  41.45 \\
 LLaVA-OV-SI-Qwen2-7b (final step)  & -      &  59.55                     &  60.30                & 41.22                      &  40.48                   &  43.31                      &  26.35                    & 45.20 \\  \midrule
Continuous Fine Tuning &  665.6      &    62.19                  &  45.76                &     43.39                  &   45.60                 &   48.11                     &    24.51                &   44.93 \\ 
Task Arithmetic(addition)~\cite{ilharco2022editing} &   13.7     &    59.19                  &     37.72             &   46.19                    &   41.96                  &   44.27                      &  26.08                  &  42.57\\    
\bottomrule
\end{tabular}}
\label{tab: online stab}
\end{table*}





\section{Conclusion}
In this paper, we introduce AbilityLens, a unified benchmark designed to assess the perception abilities of MLLMs with a focus on deeper insights beyond absolute ranking. By tracking training dynamics, AbilityLens delivers a more granular perspective on how MLLMs evolve, highlighting the development and interaction of specific abilities across varying configurations. Furthermore, we discuss some simple solutions to enhance weak abilities in MLLM, but they are only applicable in meeting one ability conflict. We hope AbilityLens will support researchers in drawing well-founded conclusions about effective strategies for advancing MLLMs.

However, our work has some limitations. First, the benchmark coverage is still limited in the scope of the domain and metrics. Second, our focus is primarily on perception abilities, with reasoning and instruction following abilities left unaddressed. We aim to address these limitations in future work.

\bibliographystyle{abbrv}
{
\small
\bibliography{main}
}

\appendix

\section{Technical Appendices and Supplementary Material}
Technical appendices with additional results, figures, graphs and proofs may be submitted with the paper submission before the full submission deadline (see above), or as a separate PDF in the ZIP file below before the supplementary material deadline. There is no page limit for the technical appendices.


\newpage
\section*{NeurIPS Paper Checklist}


\begin{enumerate}

\item {\bf Claims}
    \item[] Question: Do the main claims made in the abstract and introduction accurately reflect the paper's contributions and scope?
    \item[] Answer: \answerYes{} 
    \item[] Justification: We claim our MLLM perception benchmark contribution in the abstract.

\item {\bf Limitations}
    \item[] Question: Does the paper discuss the limitations of the work performed by the authors?
    \item[] Answer: \answerYes{} 
    \item[] Justification: We claim our limitations in the conclusion. 

\item {\bf Theory assumptions and proofs}
    \item[] Question: For each theoretical result, does the paper provide the full set of assumptions and a complete (and correct) proof?
    \item[] Answer: \answerNA{} 
    \item[] Justification: We do not have theoretical results.

    \item {\bf Experimental result reproducibility}
    \item[] Question: Does the paper fully disclose all the information needed to reproduce the main experimental results of the paper to the extent that it affects the main claims and/or conclusions of the paper (regardless of whether the code and data are provided or not)?
    \item[] Answer: \answerYes{} 
    \item[] Justification: Our data comes from public dataset and we provide code in the supplementary.

\item {\bf Open access to data and code}
    \item[] Question: Does the paper provide open access to the data and code, with sufficient instructions to faithfully reproduce the main experimental results, as described in supplemental material?
    \item[] Answer: \answerYes{} 
    \item[] Justification: We provide the data and code in the supplementary and will open-source it to the public.

\item {\bf Experimental setting/details}
    \item[] Question: Does the paper specify all the training and test details (e.g., data splits, hyperparameters, how they were chosen, type of optimizer, etc.) necessary to understand the results?
    \item[] Answer: \answerYes{} 
    \item[] Justification: We elaborate on the dataset detail and model training settings.

\item {\bf Experiment statistical significance}
    \item[] Question: Does the paper report error bars suitably and correctly defined or other appropriate information about the statistical significance of the experiments?
    \item[] Answer: \answerNo{} 
    \item[] Justification: We report the performance with trustworthy LLM eval tools.

\item {\bf Experiments compute resources}
    \item[] Question: For each experiment, does the paper provide sufficient information on the computer resources (type of compute workers, memory, time of execution) needed to reproduce the experiments?
    \item[] Answer: \answerYes{} 
    \item[] Justification: Our benchmark is efficient with only 4.5 GPU H100 GPU hours.
    
\item {\bf Code of ethics}
    \item[] Question: Does the research conducted in the paper conform, in every respect, with the NeurIPS Code of Ethics \url{https://neurips.cc/public/EthicsGuidelines}?
    \item[] Answer: \answerYes{} 
    \item[] Justification: Our benchmark meets  NeurIPS Code of Ethics.

\item {\bf Broader impacts}
    \item[] Question: Does the paper discuss both potential positive societal impacts and negative societal impacts of the work performed?
    \item[] Answer: \answerYes{} 
    \item[] Justification: We claim broader impact in the supplementary.
    
\item {\bf Safeguards}
    \item[] Question: Does the paper describe safeguards that have been put in place for responsible release of data or models that have a high risk for misuse (e.g., pretrained language models, image generators, or scraped datasets)?
    \item[] Answer: \answerNA{} 
    \item[] Justification:  Our benchmark is collected from public datasets, which are free from such cases.

\item {\bf Licenses for existing assets}
    \item[] Question: Are the creators or original owners of assets (e.g., code, data, models), used in the paper, properly credited and are the license and terms of use explicitly mentioned and properly respected?
    \item[] Answer: \answerNo{} 
    \item[] Justification: Our benchmark is collected from public datasets, which are free from such cases. We will elaborate on it on the Appendix.

\item {\bf New assets}
    \item[] Question: Are new assets introduced in the paper well documented and is the documentation provided alongside the assets?
    \item[] Answer: \answerYes{} 
    \item[] Justification: Our benchmark is well documented.

\item {\bf Crowdsourcing and research with human subjects}
    \item[] Question: For crowdsourcing experiments and research with human subjects, does the paper include the full text of instructions given to participants and screenshots, if applicable, as well as details about compensation (if any)? 
    \item[] Answer: \answerNA{} 
    \item[] Justification: This paper does not involve crowdsourcing nor research with human subjects

\item {\bf Institutional review board (IRB) approvals or equivalent for research with human subjects}
    \item[] Question: Does the paper describe potential risks incurred by study participants, whether such risks were disclosed to the subjects, and whether Institutional Review Board (IRB) approvals (or an equivalent approval/review based on the requirements of your country or institution) were obtained?
    \item[] Answer: \answerNA{} 
    \item[] Justification: This paper does not involve crowdsourcing nor research with human subjects.

\item {\bf Declaration of LLM usage}
    \item[] Question: Does the paper describe the usage of LLMs if it is an important, original, or non-standard component of the core methods in this research? Note that if the LLM is used only for writing, editing, or formatting purposes and does not impact the core methodology, scientific rigorousness, or originality of the research, declaration is not required.
    \item[] Answer: \answerNA{} 
    \item[] Justification: We use LLM for paper writing.

\end{enumerate}

\end{document}